\pdfoutput=1
\documentclass[11pt]{article}

\usepackage[final]{acl} 

\usepackage{times}
\usepackage{latexsym}
\usepackage{microtype}
\usepackage[T1]{fontenc}
\usepackage{inconsolata}
\usepackage[utf8]{inputenc}

\usepackage{xurl}
\usepackage{array}
\usepackage{amsmath}
\usepackage{tabularx}
\usepackage{enumitem}
\usepackage{ragged2e}
\usepackage{multicol}
\usepackage{multirow}
\usepackage{graphicx}
\usepackage{booktabs}
\usepackage{adjustbox}
\usepackage{subcaption}

\usepackage[compact]{titlesec}
\titlespacing{\section}{0pt}{2ex}{1ex}
\titlespacing{\subsection}{0pt}{1.5ex}{0.5ex}
\titlespacing{\subsubsection}{0pt}{1ex}{0.5ex}

\title{Figuratively Speaking: Authorship Attribution \\via Multi-Task Figurative Language Modeling}

\author{Gregorios A Katsios$^{*}$ \and Ning Sa$^{**}$ \and Tomek Strzalkowski$^{*,**}$ \\
*Department of Computer Science, **Department of Cognitive Science \\
Rensselaer Polytechnic Institute \\
\texttt{\{katsig, san2, tomek\}@rpi.edu}}

\begin{document}
\maketitle
\begin{abstract}
The identification of Figurative Language (FL) features in text is crucial for various Natural Language Processing (NLP) tasks, where understanding of the author's intended meaning and its nuances is key for successful communication. At the same time, the use of a specific blend of various FL forms most accurately reflects a writer's style, rather than the use of any single construct, such as just metaphors or irony. Thus, we postulate that FL features could play an important role in Authorship Attribution (AA) tasks. We believe that our is the first computational study of AA based on FL use. Accordingly, we propose a Multi-task Figurative Language Model (MFLM) that learns to detect multiple FL features in text at once. We demonstrate, through detailed evaluation across multiple test sets, that the our model tends to perform equally or outperform specialized binary models in FL detection. Subsequently, we evaluate the predictive capability of joint FL features towards the AA task on three datasets, observing improved AA performance through the integration of MFLM embeddings.
\end{abstract}

\section{Introduction}
Figurative Language (FL) constructs, such as metaphor, simile, and irony, are common in various forms of communication, such as literature, poetry, and speech. Their use can enrich the meaning, creativity, and persuasiveness of a message and help to achieve an intended impact on the reader. The use of certain forms of FL in writing reflects the authors' style and background, including their education, personality, social context, and worldviews. Therefore, we hypothesise that the choice of figurative language features in (written) communication may reveal the writer's cognitive and linguistic basis that underlie their production, and how their selection is influenced by the context, the intention, and the emotion of the writer. 

In this paper, we introduce a multi-task classification model designed to detect multiple Figurative Language (FL) features in a body of text. The first research question (RQ1) we seek to answer is: "Is a model that is trained to detect multiple FL features simultaneously more effective than multiple specialized models, each trained to detect a specific FL feature?" Through our research, we demonstrate that this multi-task model is indeed more effective than using several binary models.

In our research, we utilize 13 publicly available datasets to train and evaluate both binary and multi-task models. We deliberately opted against integrating additional datasets specifically designed for metaphor detection, which is only one of the phenomena we study. The rationale behind this decision was creating a more balanced training data, which otherwise would have been disproportionately skewed our study towards metaphor detection, given the substantially more resources dedicated to this phenomenon. At the same time, the lack of annotated corpora for other figurative language features such as personification, metonymy, oxymoron, etc. necessarily limited our initial study to the six FL constructs that are generally well represented among these 13 datasets: Metaphor, Simile, Idiom, Sarcasm, Hyperbole, and Irony.

All our binary models and the multi-task model are based on RoBERTa \cite{liu2019roberta}. After training the specialized binary models on the combined datasets, we used them to automatically label our training corpora with all applicable FL features. This multi-label dataset was then used to train our multi-task model. Afterwards, we compare our Multi-task Figurative Language Model (MFLM) against the binary classifiers on the 13 test sets. The results showed that MFLM matched or outperformed the binary classifiers in five test sets and achieved higher task-specific performance than the binary models in another three test sets, which suggests that these features are not independent from one another. 

After training our multi-task figurative language classifier, we put forward a second research question (RQ2): "Does the incorporation of Figurative Language (FL) features enhance performance in Authorship Attribution (AA) tasks?" To answer this, we evaluate the impact of the FL features learned by our Multi-task Figurative Language Model (MFLM) on three publicly available AA datasets, each consisting of documents with varying topical content and number of authors. For each dataset, we train Multi-Layer Perceptron (MLP) classifiers, using MFLM sentence embeddings and other baselines as input features. The baselines consist of classical Stylometric features, character and word $n$-gram TF-IDF vectors, and generic sentence embeddings. Our results demonstrate that the AA task performance is indeed improved by combining MFLM embeddings with other baselines. 


To our knowledge, this work is the first to examine the applicability of FL features in AA. We should note here that we did not expect that the FL features alone would be sufficient to perform AA; rather we set off to demonstrate that incorporating combined FL features improves AA performance when integrated with more basic stylistic features, particularity for longer texts. The results show that the latter is generally true; however, we found that the FL features perform nearly as strong and sometimes better on their own. This supports our initial stipulation that FL use is highly personalized, and thus an excellent predictor of authorship.

We make our code and data available in our GitHub repository\footnote{Figuratively Speaking: \url{https://github.com/HiyaToki/Figuratively-Speaking}}.

\section{Related Work}
Most of the previous studies on Figurative Language (FL) feature detection focus on the features independently. An earlier work, \citep{tsvetkov2014metaphor}, used lexical semantic features of the words to discriminate metaphors from literals. More recently, \citet{choi2021melbert} utilized metaphor identification theories using RoBERTa to predict whether a word in a sentence is metaphorical or not. A similar shift from linguistic feature based approach to pre-trained language model (PLM) based approach is observed in simile detection. \citet{niculae2014brighter} extracted features such as topic-vehicle similarity and imageability to separate similes from literal comparisons. \citet{ma2023run} used BERT \citep{devlin2018bert} and RoBERTa in simile property probing tasks and concluded that the PLMs still underperformed humans. PLMs are also applied to the detection of sarcasm \citep{yuan2022sarcasm}, hyperbole \citep{biddle2021hypo}, irony \citep{gonzalez2020irony}, and idiom \citep{Briskilal2022idiom}.

Among the studies that work on more than one features, \citet{badathala2023match} used datasets cross-labeled with metaphor and hyperbole, and found that the multi-task learning approach performed better than the single-task approach on both features. \citet{chakrabarty2022flute} rendered the FL detection into a multi-task natural language inference (NLI) problem, developed a NLI dataset of four FL features, and tested with several experimental systems. \citet{chakrabarty2022simile} collected datasets on idiom and simile and developed knowledge enhanced RoBERTa-based models. However, their task was to predict the correct continuation of the given narrative, not FL feature detection. \citet{adewumi2021potential} built a dataset covering 9 FL features plus literals. They tested three baseline systems in a multi-class classification task and BERT outperformed the other two systems. 

There is a rich literature in the field of Authorship Attribution (AA). Various methods have been applied to the task, ranging from SVM based approaches, such as \citep{kestemont2018overview}, to transformer based models, like \citep{Bauersfeld2023aa}. In PAN-2019 cross-domain AA challenge \citep{kestemont2019pan}, most of the submissions used $n$-gram features (char, word, part-of-speech) and an ensemble of classifiers (SVM, Logistic Regression, etc). \citet{fabien2020bertaa} fine-tuned a BERT model for AA task and tested the model on three datasets including IMDB-62 \citep{seroussi2014authorship}. In a recent review article \cite{tyo2022aa-review}, feature based methods and embedding based methods were tested and compared on the same datasets. They used $n$-grams, summary statistics and co-occurance graphs as features, as well as static char/word embeddings and transformer-based sentence embeddings. 

\section{Figurative Language Modeling}
In our study, we investigate the potential benefits of combining Figurative Language (FL) features as opposed to analyzing each feature independently. To answer our first research question, we examine whether training a FL classification model capable of jointly labeling text with relevant features would outperform a singular binary model specialized in detecting only one feature. This idea stems from noticing that in both spoken and written language, individuals intertwine various elements of figurative speech to effectively convey their intended message. Consequently, FL features frequently co-occur, and understanding the interplay between these features may offer valuable insights for improving their identification accuracy. This research builds upon prior studies that explored the simultaneous detection of metaphors and sarcasm, as well as hyperbole and sarcasm. In our investigation, we aim to simultaneously learn to detect six distinct FL features: Metaphors, Simile, Sarcasm, Hyperbole, Idiom, and Irony. 

\begin{figure*}[hbt]
    \centering
    \includegraphics[width=0.85\textwidth]{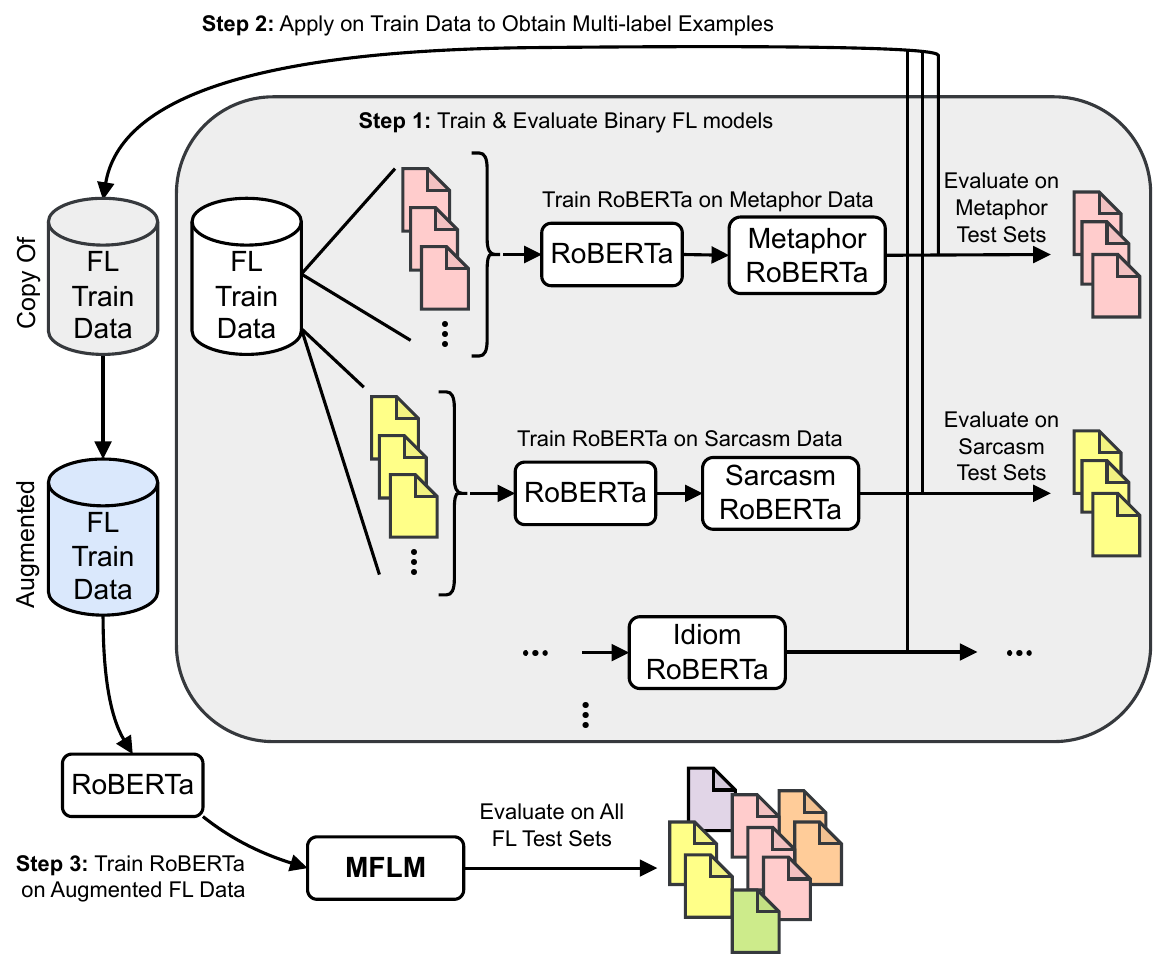}
    \caption{Diagram illustrating our pipeline of training the individual binary FL models, augmenting the FL training collection with predicted labels and fine-tuning the MFLM.}
    \label{fig:pipeline}
\end{figure*}

\subsection{Data} \label{sec:data}
In our research to learn to classify FL phenomena, we rely on publicly available datasets. In total, we work with 13 individual corpora, which are summarized in Table \ref{tab:datasets} (see Appendix \ref{sec:appendix} for additional details). While space constraints prevent exhaustive descriptions, we encourage interested readers to explore the original works by the dataset creators for comprehensive insights into the data collection and annotation processes.

Among the datasets we analyze, the iSarcasm corpus \cite{farha2022semeval} stands out as truly multi-labeled. It includes training and testing examples annotated with labels such as sarcasm, irony, overstatement (hyperbole), understatement, satire, and rhetorical questions. For instance, an excerpt from the iSarcasm training set reads: \textit{"Can’t wait to be back at uni so I can order more shoes and clothes without my mum telling me off"}, which is labeled with both sarcasm and hyperbole.

In contrast, several other datasets adopt a multi-class approach. Each example in these datasets corresponds to a single applicable label. Additionally, some datasets focus exclusively on specific FL phenomena, employing positive and negative examples (e.g., \textit{feature\_X} and \textit{not\_feature\_X}) to create a binary distinction.

When dealing with FL datasets, it’s crucial to consider how negative examples are constructed. Some datasets construct the negative class (i.e., \textit{not\_feature\_X}) by ensuring that samples represent true literal sentences devoid of any FL speech. The FLUTE corpus \cite{chakrabarty2022flute} is an example of this approach, where FL sentences are paired with their rephrased literal counterparts. For instance, the figurative sentence (metaphor): \textit{"A break up can leave you with a broken heart"} is paired with the literal sentence: \textit{"It's hurtful when a breakup makes you feel lonely and sad"}.

Other datasets annotate the negative class as simply not containing the FL phenomena described by the positive class. For instance, in the Irony SemEval 2018 corpus \cite{van2018semeval}, sentences that are labeled as \textit{not\_irony} may still exhibit other FL traits. Consider the sentence: \textit{"Look for the girl with the broken smile"} which, although not ironic, contains a metaphor that is not explicitly annotated. 

In our pipeline, we apply minimal pre-processing to the sentences from these corpora, and we load them into our combined collection, retaining human annotations relevant to our work. Notably, we focus on the six FL features listed in Table \ref{tab:datasets}, ignoring classes beyond this scope. At this stage, we clearly distinguish between literal sentences and negative class sentences labeled as \textit{not\_feature\_X}.

In our study, we encounter various datasets with distinct characteristics regarding their train/dev/test splits. Some datasets come with a predefined splits, where we merge the training and development sets into a single training set, reserving the original test set solely for evaluation. In cases where datasets lack existing splits, we adopt a systematic approach, setting aside a 10\% stratified sample for testing. The entire collection consists of 69168 training and 9729 testing examples.

\begin{table*}[htb]
  \centering
  \small
  \begin{tabular}{>{\raggedright}p{0.33\linewidth}|c|c|c|c|c|c|c}
    \hline
    \textbf{Datasets} & \textbf{Metaphor} & \textbf{Simile} & \textbf{Sarcasm} & \textbf{Hyperbole} & \textbf{Idiom} & \textbf{Irony} & Literal \\
    \hline
    Reddit Irony Corpus \citep{wallace2014humans} & - & - & - & - & - & 537 & No \\
    \hline
    Irony SemEval18 \citep{van2018semeval}        & - & - & - & - & - & 2212 & No \\
    \hline
    iSarcasm \citep{farha2022semeval}             & - & - & 846 & 46 & - & 174 & No \\
    \hline
    Sarcasm Corpus \citep{oraby2017creating}      & - & - & 4693 & 1164 & - & - & No \\
    \hline
    MOVER \citep{zhang2021mover}                  & - & - & - & 1007 & - & - & Yes \\
    \hline
    HypoGen \citep{tian2021hypogen}               & - & - & - & 1876 & - & - & Yes \\
    \hline
    EPIE \citep{saxena2020epie}                   & - & - & - & - & 2761 & - & Yes \\
    \hline
    PIE-English \citep{adewumi2021potential}      & 12590 & 1072 & 46 & - & 13738 & 30 & Yes \\
    \hline
    FLUTE \citep{chakrabarty2022flute}            & 749 & 750 & 2677 & - & 1009 & - & Yes \\
    \hline
    MSD23 \citep{ma2023run}                       & - & 3576 & - & - & - & - & Yes \\
    \hline
    Figurative Comparisons \citep{niculae2014brighter} & - & 449 & - & - & - & - & Yes \\
    \hline
    LCC \citep{mohler2016introducing}             & 3036 & - & - & - & - & - & No \\
    \hline
    MOH \citep{mohammad2016metaphor}              & 410 & - & - & - & - & - & Yes \\
    \hline
  \end{tabular}
  \caption{Datasets used in our multi-task Figurative Language approach. The values in the cells denote the number of examples per feature. The last column indicates whether a dataset employs literal examples to form the negative class. Blank fields correspond to datasets (rows) that do not contain any annotations for the corresponding FL feature (columns).}
  \label{tab:datasets}
\end{table*}

\subsection{Binary Models}
To detect the various FL phenomena, we create task-specific binary classifiers. This process involves combining datasets annotated with examples relevant to each specific feature. For instance, to train a classifier for metaphors, we aggregate data from PIE-English, FLUTE, LCC, and MOH datasets. Similarly, for simile classification, we gather data from PIE-English, FLUTE, MSD23, and Figurative Comparisons datasets.

The combination of datasets allows us to establish both positive and negative sets for each classification task. In the context of training a metaphor classifier, the positive set comprises examples exhibiting metaphoric expressions, while the negative set encompasses instances without metaphors. As detailed in Section \ref{sec:data}, certain datasets exclusively utilize literal examples for constructing the negative class, whereas others use examples not containing the FL phenomena described by the positive class. Thus, achieving a balanced representation necessitates the inclusion of negative samples from both types of datasets.

In our approach, positive and negative examples are retrieved from the combination datasets corresponding to the specific task, while literal examples are sourced from across all datasets. The final training set for each task is formed by selecting all positive examples and supplementing them with an equal number of negative and literal examples. Specifically, if the size of the positive class is denoted as $N$, we sample $N/2$ negative and $N/2$ literal examples. In scenarios where there are insufficient negative examples, we augment the dataset with an appropriate number of literal examples to ensure a total of $2N$ training instances. During training, the labels of literal examples are transformed to \textit{not\_feature\_X}, aligning with our objective to create robust binary classifiers capable of discerning sentences containing the specific feature from those that do not. For detailed information on the number of training samples per task, please refer to the Appendix \ref{sec:appendix}.

Subsequently, we train individual RoBERTa \cite{liu2019roberta} models\footnote{RoBERTa-Large: \url{https://huggingface.co/FacebookAI/roberta-large}} for each task using a standardized set of hyper-parameters across all training jobs: Epochs: 5, Learning Rate: 2e-5,  Weight Decay: 0.01, Warm-up Ratio: 0.1, Batch Size: 16. The time required to train a binary model averaged at approximately 80 minutes, using a single NVidia RTX A6000 GPU.

\subsection{Multi-Task Model}
We proceed to train a Multi-task Figurative Language Model (MFLM) that can label a sentence with all applicable features in a single pass. For this, we convert our combined training dataset into a multi-label format. We use the array of binary models to assign all the possible labels to each training sentence in our corpora. 


Consequently, we obtain an augmented FL training corpus, for which  every sentence has a corresponding list of predicted FL labels. To produce a high quality training set, we keep only the examples where the predicted labels are consistent with the original human annotations. For instance, if a sentence is annotated by humans as: \textit{[metaphor, idiom]}, we accept predictions such as: \textit{[metaphor, idiom, simile, not\_irony, not\_hyperbole, not\_sarcasm]}, but we reject predictions like: \textit{[metaphor, not\_idiom, not\_simile, not\_irony, not\_hyperbole, not\_sarcasm]}, due to the \textit{not\_idiom} prediction's inconsistency.

In this manner we create a dataset of 61264 sentences, discarding 7904 text-prediction pairs that conflict with human annotations. The distribution of labels in the dataset is shown in Table \ref{tab:join_dataset}. We allocate 10\% of this training set to be used as a development set, facilitating the identification of the optimal probability threshold for each feature. Leveraging both automatically generated labels and human annotations, we obtain two distinct sets of thresholds. One set is optimized based on the human labels, while the other set is calibrated using the automatic labels.

\begin{table}[htb]
  \centering
  \small
  \begin{tabular}{c|c}
    \hline
    \textbf{Metaphor} & 18981 \\
    \hline
    \textbf{Simile} & 6618 \\
    \hline
    \textbf{Sarcasm} & 9906 \\
    \hline
    \textbf{Hyperbole} & 13699 \\
    \hline
    \textbf{Idiom} & 18604 \\
    \hline
    \textbf{Irony} & 10176\\
    \hline
  \end{tabular}
  \caption{Class distribution of the automatically annotated multi-label training dataset.}
  \label{tab:join_dataset}
\end{table}

We follow the same hyper-parameter set-up as the binary model training, and the average time to train the multi-task model is about 326 minutes, using a single NVidia RTX A6000 GPU. Our pipeline of training the individual binary FL models, augmenting the FL training collection with predicted labels and fine-tuning the MFLM, is illustrated in Figure \ref{fig:pipeline}.

\subsection{Evaluation and Results}
To evaluate both binary and multi-task approaches, we use the reserved task-specific testing sets. In Tables \ref{tab:multi_results_1} and \ref{tab:multi_results_2}, we report the weighted average F1-score obtained from a single run. The rows marked as Metaphor, Simile, Sarcasm, Hyperbole, Idiom and Irony refer to binary models while the rows marked as MFLM refer to our multi-task model. MFLM-h and MFLM-b refer to predictions acquired by tuning the probability thresholds on the development set using human annotations and binary predictions respectively. 

Due to space limitations, we present a single column for the multi-class test sets. Nonetheless, our binary models were evaluated appropriately, by treating  annotations from unrelated tasks as \textit{not\_feature\_X}. For instance, when evaluating the Metaphor binary model on the FLUTE test set, simile, sarcasm and idiom ground truth labels become \textit{not\_metaphor}. In contrast, since our MFLM can inherently support all classes, we report weighted F1-score without altering the ground truth labels.

\begin{table*}[htb]
  \begin{subtable}{\linewidth}
    \centering
    \small
    \begin{tabular}{c|c|c|c|c|c|c}
        \hline
                           & FLUTE & iSarcasm & Sarcasm Corpus & MSD23 & Figurative Comparisons & LCC \\
        \hline
        \textbf{Metaphor}  & 0.76 & -    & -    & -    & -    & 0.83 \\
        \hline
        \textbf{Simile}    & 0.98 & -    & -    & 0.80 & 0.81 & -    \\
        \hline
        \textbf{Sarcasm}   & 0.97 & 0.82 & 0.80 & -    & -    & -    \\
        \hline
        \textbf{Hyperbole} & -    & 0.96 & 0.56 & -    & -    & -    \\
        \hline
        \textbf{Idiom}     & 0.85 & -    & -    & -    & -    & -    \\
        \hline
        \textbf{Irony}     & -    & 0.79 & -    & -    & -    & -    \\
        \hline
        \hline
        \textbf{MFLM-h}    & 0.87 & 0.43 & \underline{0.70} & \textbf{0.84} & 0.78 & 0.67 \\
        \hline
        \textbf{MFLM-b}    & \underline{0.89} & 0.37 & \underline{0.72} & \textbf{0.83} & \textbf{0.81} & 0.74 \\
        \hline
    \end{tabular}
  \caption{Part 1 of the evaluation results.}
  \label{tab:multi_results_1}
  \end{subtable}\vspace{2ex}

  \begin{subtable}{\linewidth}
    \centering
    \small
    \begin{tabular}{c|c|c|c|c|c|c|c}
        \hline
                           & MOH & EPIE & PIE-English & Irony SemEval18 & Reddit Irony & HypoGen & MOVER \\
        \hline
        \textbf{Metaphor}  & 0.81 & -    & 0.92 & -    & -    & -    & -    \\
        \hline
        \textbf{Simile}    & -    & -    & 0.98 & -    & -    & -    & -    \\
        \hline
        \textbf{Sarcasm}   & -    & -    & -    & -    & -    & -    & -    \\
        \hline
        \textbf{Hyperbole} & -    & -    & 0.86 & -    & -    & 0.70 & 0.71 \\
        \hline
        \textbf{Idiom}     & -    & 0.91 & 0.99 & -    & -    & -    & -    \\
        \hline
        \textbf{Irony}     & -    & -    & 0.97 & 0.67 & 0.66 & -    & -    \\
        \hline
        \hline
        \textbf{MFLM-h}    & 0.58 & \textbf{0.91} & \underline{0.97} & \textbf{0.76} & 0.49 & 0.69 & 0.64 \\
        \hline
        \textbf{MFLM-b}    & 0.58 & \textbf{0.94} & \underline{0.97} & \textbf{0.71} & 0.52 & \textbf{0.77} & 0.64 \\
        \hline
    \end{tabular}
  \caption{Part 2 of the evaluation results.}
  \label{tab:multi_results_2}
  \end{subtable}\vspace{-1ex}
  
  \caption{Evaluation results on task-specific test sets. We report the weighted F1-score. With bold we draw attention to evaluation results where the MFLM is on par or surpasses the corresponding binary model. Scores that are underlined correspond to cases where the MFLM is on par or outperforms a binary model on a specific task. Blank fields correspond to binary models (rows) that are not applicable to the corresponding test set (columns).}
\end{table*}

\begin{table*}[htb]
  \centering
  \small
  \begin{tabular}{c|c|c}
    \multicolumn{3}{c}{MFLM \& Binary models disagree with GT} \\
    \hline
    \textbf{GT}   & Literal &  \multirow{3}{*}{\begin{tabular}{p{0.6\linewidth}} \textit{The guests \textbf{showered rice} on the couple.} \end{tabular}} \\
    \cline{1-2}
    \textbf{MFLM} & Metaphor & \\
    \cline{1-2}
    \textbf{Bin}  & Metaphor & \\
    \hline
    \hline    
    \textbf{GT}   & Literal &  \multirow{3}{*}{\begin{tabular}{p{0.6\linewidth}} \textit{Charlie'd asked me if I'd like to \textbf{make a bit on the side}.} \end{tabular}} \\
    \cline{1-2}
    \textbf{MFLM} & Metaphor, Idiom & \\
    \cline{1-2}
    \textbf{Bin}  & Metaphor & \\
    \hline
    \hline
    \textbf{GT}   & Not Irony &  \multirow{3}{*}{\begin{tabular}{p{0.6\linewidth}} \textit{And \textbf{seeing the light on the current drug policies}.} \end{tabular}} \\
    \cline{1-2}
    \textbf{MFLM} & Metaphor, Idiom, Irony & \\
    \cline{1-2}
    \textbf{Bin}  & Metaphor, Idiom, Irony & \\
    
    \hline
    \multicolumn{3}{c}{ } \\
    \multicolumn{3}{c}{MFLM disagrees with GT, Binary models agree with GT} \\
    \hline    
    \textbf{GT}   & Literal &  \multirow{2}{*}{\begin{tabular}{p{0.6\linewidth}} \textit{She said he was very nice and he \textbf{beamed a smile} at her.} \end{tabular}} \\
    \cline{1-2}
    \textbf{MFLM} & Metaphor & \\
    \hline
    \hline
    \textbf{GT}   & Literal &  \multirow{2}{*}{\begin{tabular}{p{0.6\linewidth}} \textit{I was talking to someone and \textbf{we had great chemistry} then they \textbf{went ghost on me}.} \end{tabular}} \\
    \cline{1-2}
    \textbf{MFLM} & Idiom & \\
    \hline
    \hline
    \textbf{GT}   & Literal &  \multirow{2}{*}{\begin{tabular}{p{0.6\linewidth}} \textit{I can't get pregnant but all my friends are having kids.} \end{tabular}} \\
    \cline{1-2}
    \textbf{MFLM} & Irony & \\
    \hline

  \end{tabular}
  \caption{Error analysis - samples where model predictions do not align with human annotations (ground truth).}
  \label{tab:multi_analysis}
\end{table*}

The MFLM demonstrates competitive or superior performance compared to binary classifiers across different test sets. Specifically, in 5 out of 13 tests, the MFLM either matches or surpasses binary models. Furthermore, in 3 tests, the MFLM exhibits comparable or superior performance in specific tasks. For instance, MFLM-h performs equally well as the Simile and Sarcasm models on the FLUTE test sets, achieving F1-scores of 0.98 and 0.97 respectively. Moreover, the MFLM-h surpasses the Sarcasm model on the Sarcasm Corpus test set, with F1-scores of 0.82 and 0.80 respectively. On the same test set, the Hyperbole model outperforms the MFLM-h in the hyperbole task, with F1-scores of 0.56 and 0.33 respectively. In the PIE-English test set, the MFLM-h excels over the Metaphor binary model on the metaphor task with 0.96 versus 0.92 F1-score respectively, and matches the performance of the Idiom model. This supports our first research question and highlights the versatility and effectiveness of the MFLM across different linguistic tasks and datasets.

\subsubsection{Error Analysis}
To pinpoint the weaknesses and strengths of our MFLM, we conduct a manual error analysis, scrutinizing samples where the multi-task and/or binary models disagree with the ground truth. For each case, we display a few random examples in Table \ref{tab:multi_analysis}, while more samples are presented in the Appendix \ref{sec:appendix} for further reference. Our findings indicate that the majority of miss-classifications made by the MFLM stem from inaccuracies or incompleteness in the annotation of input sentences. Nonetheless, the predictions generated by the MFLM demonstrate a reasonable level of accuracy in most instances and carry on to experiment using our proposed multi-task FL model and evaluate its appropriateness on the Authorship Attribution (AA) task.

\section{Authorship Attribution}
We proceed to investigate the effectiveness of our Multi-task Figurative Language Model (MFLM) in the closed-case Authorship Attribution (AA) downstream task. AA involves classifying texts to determine their respective authors from a known set of candidates. Specifically, given a training corpus consisting of $N$ authors, the objective is to predict the author of each document in the test set by selecting from the set of $N$ authors.

Our second research question proposes that embeddings incorporating figurative language features will enhance performance in the AA task. This concept extends from stylometric analysis \citep{lagutina2019stylometric}, which traditionally concentrates on discerning patterns within written text. Stylometric analysis examines various aspects of writing style, including word selection, sentence construction, punctuation usage, and vocabulary preferences. To the best of our knowledge, our study is the first of its kind to utilize a Transformer model that has been fine-tuned for multi-task FL classification, towards the AA task. Previous research in this area minimally explored the applicability of FL features for this specific task.

\begin{figure*}[htb]
    \centering
    \includegraphics[width=0.85\textwidth]{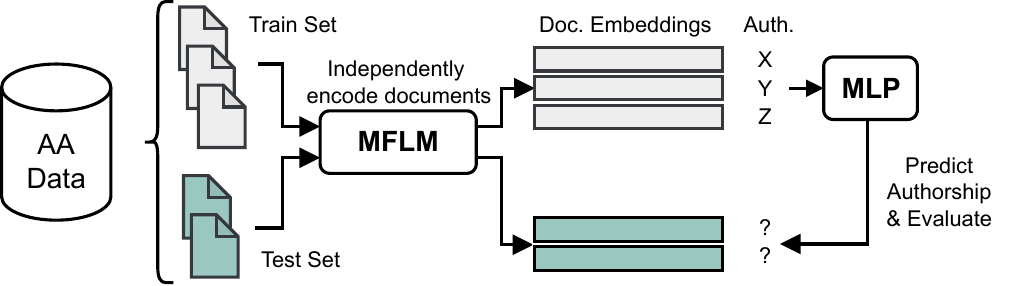}
    \caption{Diagrammatic representation of our Authorship Attribution training and evaluation approach. Following this process, any baseline can take the place of the "MFLM" rectangle.}
    \label{fig:aa_pipe}
\end{figure*}

\subsection{Data}
In our Authorship Attribution (AA) experiments, we employ three distinct, publicly accessible datasets. The first dataset, IMDb-62 \citep{seroussi2014authorship}, comprises 1000 movie reviews from each of the 62 authors. These reviews are relatively short, averaging around 100 words. The IMDB-62 dataset does not have a predetermined train/test split, therefore we reserve a 10\% stratified sample for testing. This yields a training set of 55800 examples and a testing set of 6200 samples.

The second dataset, PAN-2006 \citep{houvardas2006n}, is focused on corporate and industrial topics. It includes short texts of approximately 500 words. The training set comprises 2500 texts, with 50 texts per author. Similarly, the test set consists of 2500 texts, with 50 texts per author, ensuring no overlap with the training data.

The third and final dataset, PAN-2018 \citep{kestemont2018overview}, contains medium-length texts of around 800 words each, centered on fan fiction. This dataset is divided into four problems, each with a different number of authors (20, 15, 10, and 5). However, each author consistently contributes seven texts. The test sets vary in the number of texts they contain, with 79, 74, 40, and 16 texts respectively. In our experiments, we use only the English texts.

\subsection{Baselines}
In our Authorship Attribution (AA) task, we evaluate the performance of our MFLM against four different baselines. The first baseline is built upon classical Stylometric features. We implement 52 text metrics using the cophi\footnote{cophi: \url{https://github.com/cophi-wue/cophi-toolbox}} and textstat\footnote{textstat: \url{https://github.com/textstat/textstat}} Python packages. These metrics are used to form a document vector with 52 stylometric features. For a more detailed explanation of these features, please refer to the Appendix \ref{sec:appendix}.

The second baseline utilizes the all-roberta-large-v1\footnote{all-roberta-large-v1: \url{https://huggingface.co/sentence-transformers/all-roberta-large-v1}} Sentence Embedding \citep{reimers2019sentence} model, which we refer to as SBERT in the following sections. This model is comparable to our MFLM since it is also based on RoBERTa-Large, but without the multi-task FL classification fine-tuning. With SBERT, we generate a 1024-dimensional document vector. This vector is computed by averaging the individual sentence embeddings for each input text.

The third and fourth baselines in our study are constructed using word and character $n$-grams, respectively. We utilize the Python package scikit-learn \cite{scikit-learn} to analyze the texts and identify the 1024 most common $n$-grams from the training dataset, where the value of $n$ varies from 1 to 5. We exclude stop words from the input texts during this process. Subsequently, we compute the Term Frequency-Inverse Document Frequency (TF-IDF) values for these $n$-grams across all documents, resulting in 1024-dimensional sparse document vectors.

\subsection{Evaluation and Results}
For the evaluation, we begin by encoding all texts in the AA datasets utilizing our MFLM model and the baselines. To create the embeddings using the MFLM, we discard the multi-task classification layer and directly utilize the underlying Transformer model. The sentence embedding is computed by mean-pooling all token embeddings, including the \textit{[CLS]} token, taken from the last hidden layer. To create the document embedding, we average the embeddings of individual sentences. This allows us to create a 768-dimensional vector for each document. 

Following this encoding step, we construct Multi-Layer Perceptron (MLP) classifiers for each test case and features combination. These MLP models consist of a single hidden layer comprising 1024 units and are implemented using the Python package scikit-learn. Our training process involves 1000 epochs with a learning rate of 2e-5, incorporating early stopping. The activation function employed is ReLU \cite{agarap2018deep}, and the optimizer used is Adam \cite{kingma2014adam}. Subsequently, we apply the trained model on the test set to calculate weighted average F1-scores obtained from a single run, which are presented in Table \ref{tab:aa_results}. The training and evaluation process for Authorship Attribution (AA) is illustrated in Figure \ref{fig:aa_pipe}.

\begin{table*}[!htb]
  \centering
  \small
  \begin{tabular}{c|c|c|c|c|c|c||c|c|c||c}
    \textbf{\#Auth} & \textbf{Dataset} & MFLM & Stylo & SBERT & Word & Char & SBERT+ & Word+ & Char+ & SOTA \\ \hline
    62 & IMDB62   & 0.91 & 0.02 & 0.87 & 0.82 & \textbf{0.92} & 0.94 & 0.96 & 0.96 & 0.99$^*$ \\ \hline
    50 & PAN06    & 0.58 & 0.00 & 0.62 & \textbf{0.65} & 0.64 & 0.64 & 0.68 & 0.66 & 0.77$^*$ \\ \hline
    20 & PAN18 P1 & \textbf{0.57} & 0.01 & 0.30 & 0.40 & 0.52 & 0.55 & 0.59 & 0.60 & 0.65$^{**}$ \\ \hline
    15 & PAN18 P2 & \textbf{0.63} & 0.00 & 0.40 & 0.50 & 0.62 & 0.64 & 0.66 & 0.66 & 0.68$^{**}$ \\ \hline
    10 & PAN18 P3 & 0.67 & 0.20 & 0.50 & 0.63 & \textbf{0.82} & 0.67 & 0.71 & 0.74 & 0.74$^{**}$ \\ \hline
    5  & PAN18 P4 & 0.57 & 0.20 & 0.49 & \textbf{0.69} & 0.57 & 0.64 & 0.69 & 0.66 & 0.68$^{**}$ \\ \hline
  \end{tabular}
  \caption{Authorship Attribution evaluation results, reporting the weighted F1-score. With bold font, we draw attention to the best performing model. Word and Char refer to the word $n$-grams and character $n$-grams respectively. Columns where the title line contains a `+' character correspond to experiments where we concatenated MFLM embeddings to the baseline document vector. $^*$Macro-averaged accuracy as reported in \citep{tyo2022aa-review}. $^{**}$Macro-averaged F1-score as reported in \citep{kestemont2018overview}.}
  \label{tab:aa_results}
\end{table*}

Character and word $n$-grams features remain a valuable tool for AA, as their strength lies in capturing stylistic features like word choice, punctuation, and common phrases, often unique to an author. $N$-gram features, encompassing character sequences, spelling preferences, and even made-up words, remain consistent even with smaller datasets and paraphrasing. This robustness makes them effective for identifying rare words, misspellings, and author-specific quirks. However, they lack the ability to capture the semantic and pragmatic aspects of meaning or structural organization of text (which we do not address in this paper), both essential aspects of an author's overall style.

On the other hand, MFLM document vectors address both semantic and pragmatic aspects by encoding Figurative Language (FL) features within sentences. This approach allows for a more nuanced comparison of texts, considering not only the use of metaphors, similes, and other rhetorical devices by the author, but also their unique combinations. This could potentially lead to a more effective generalization across various writing styles and genres. Prior work on FL and metaphors \cite{lakoff2008metaphors,thibodeau2009bad} has noted that authors often blend their FL constructs in a seemingly haphazard manner. Rather than conforming to any discernible "logic", this pattern seems to be a reflection of the author's individual style, as suggested by our findings. 
While quite powerful, FL-based features don't encompass all facets of an individual's writing style. We continue to investigate the structural aspects of texts, which is one area that remains under study. On the other end of the spectrum, we must also account for information contained in subword patterns, an area where $n$-grams excel. Additionally, typos, grammatical errors, and paraphrasing can significantly impact MFLM embeddings, potentially resulting to misleading attributions.

Furthermore, we conducted experiments to explore the impact of integrating Figurative Language (FL) features by combining our MFLM encoding with baseline document vectors and subsequently training new MLP classifiers. Our findings demonstrate a consistent boost in performance across nearly all cases when using the combined features, thereby supporting our second research question.

In Table \ref{tab:aa_results}, we also include state-of-the-art (SOTA) results, as reported in \citep{tyo2022aa-review} and \citep{kestemont2018overview}. The methodologies vary across implementations, but character $n$-grams, part-of-speech $n$-grams, and summary statistics typically form the input for an ensemble of logistic regression classifiers, achieving SOTA in the AA task.  It is important to note that in \citep{tyo2022aa-review}, the authors report macro-averaged accuracy, while in \citep{kestemont2018overview}, the evaluation metric is macro-averaged F1-score. Although a direct comparison may not be feasible due to these differing metrics, these results offer valuable insight into the task's complexity.

\section{Conclusion}
This study investigated two research questions regarding the detection and application of Figurative Language (FL) features in machine learning.

Firstly, we explored whether a multi-task model trained to simultaneously detect multiple FL features (Metaphor, Simile, Idiom, Sarcasm, Hyperbole, and Irony) could outperform individual models specialized for each feature. By leveraging RoBERTa-Large and a multi-label training dataset derived from binary classifiers, our Multi-task Figurative Language Model (MFLM) achieved superior performance on 8 out of 13 test sets, particularly excelling in detecting Simile, Idiom, Irony, and Hyperbole. This finding highlights the increased effectiveness of a unified approach for comprehensive FL detection.

Secondly, we examined the potential of incorporating FL features to enhance performance in Authorship Attribution (AA) tasks. Utilizing three diverse AA datasets and Multi-Layer Perceptron (MLP) classifiers, we evaluated the contribution of MFLM sentence embeddings alongside various baseline features like Stylometric features, SBERT Embdeddings and word and character $n$-gram vectors. The results showed the competitive performance achieved by MFLM embeddings alone, while their combination with other features yielded consistent performance improvements across nearly all cases. This strongly supports the second research question, indicating the positive impact of integrating FL features in AA tasks.

Our study offers valuable insights into the effectiveness of multi-task learning for comprehensive FL detection and the potential of FL features to improve AA tasks. Further research could explore the applicability of MFLM to additional NLP tasks, such as sentiment analysis and information retrieval. Moreover, future studies could investigate the impact of incorporating additional FL features into a single classification model, such as personification, metonymy, onnomatopoeia, etc., as well as domain-specific knowledge for even more refined FL detection and application.

\section*{Acknowledgements}
This research is supported in part by the Office of the Director of National Intelligence (ODNI), Intelligence Advanced Research Projects Activity (IARPA), via the HIATUS Program contract \#2022-22072200002 and the Defense Advanced Research Projects Agency (DARPA) under Contract No. HR001121C0186. The views and conclusions contained herein are those of the authors and should not be interpreted as necessarily representing the official policies, either expressed or implied, of ODNI, IARPA, DARPA, or the U.S. Government. The U.S. Government is authorized to reproduce and distribute reprints for governmental purposes notwithstanding any copyright annotation therein.

\section*{Ethical Considerations and Limitations}
In this paper, we investigate the efficacy of training a multi-task classification model to detect Figurative Language (FL) features compared to specialized binary models. In this work, we also explore leveraging the multi-task model embeddings for the Authorship Attribution (AA) task.

One potential limitation of our study arises from the combination of different datasets for the various Figurative Language (FL) features under consideration. The quality of annotations across these datasets is not uniform, with some lacking annotation manuals or relying on automatic and crowd-sourced approaches for dataset creation. This inconsistency can introduce errors into our model. Furthermore, the datasets, while publicly available, may contain inherent biases due to the lack of clear instructions for annotating literal sentences and potential variability in human annotator judgments.

In the process of constructing annotated corpora for training machine learning algorithms for automatic figurative language detection, it's crucial to consider the interpretive discrepancies between experts and non-experts. The annotations found in the collection of datasets used in this study are all taken as ground truth of equal importance, potentially leading towards a biased FL detection model. An expert, with their nuanced understanding, can identify subtle metaphors and idioms that may elude an ordinary reader. However, non-experts, influenced by their unique cultural backgrounds and personal experiences, may interpret figurative language differently \citep{carrol2020resolving,robo2020discrepancies}. For instance, certain phrases may have specific connotations in one culture but be meaningless in another. Similarly, a person's familiarity with a subject matter can greatly influence their understanding of related figurative language. Therefore, to ensure a more accurate and comprehensive analysis of figurative language, these factors must be taken into account. Future work will address this issue by conducting qualitative and quantitative analyses on the annotated datasets.

In our methodology, we employ specialized binary models for each feature, trained on our combined datasets, to predict figurative language labels for the training examples used in our multi-task model. This approach, while effective, can lead to error propagation, resulting in incorrect predictions from our model. However, our evaluation and manual error analysis indicate that our multi-task model's predictions are often reasonable, with errors frequently attributable to incomplete human annotations.

The second part of our study applies the embeddings from our multi-task FL model to the AA task. We train MLP classifiers using document vectors as features on three publicly available datasets, each focusing on a different topic: movie reviews, corporate/industrial topics, and fan fiction. However, these topics are not very diverse, which could introduce bias into the datasets with respect to authorship. For instance, in the fan fiction dataset, some authors may exclusively write "Harry Potter" fan fiction, which could skew the evaluation of different features.

Lastly, it is important to note that the predictions of deep neural language models, such as the ones used in our study, are often difficult to interpret and explain. This lack of interpretability is a common challenge in the field and is another limitation to consider in our work.

\bibliography{references} 

\appendix
\section{Appendix}
\label{sec:appendix}

\subsection{Appendix: Figurative Language Datasets}
Here we present additional details regarding our 13 figurative language datasets. In Tables \ref{tab:multi_dataset_distribution_1} and \ref{tab:multi_dataset_distribution_2} we show the number of examples per class label for the train/test sets of all datasets. The datasets that had predefined train/test splits are: FLUE, iSarcasm, and Irony SemEval 2018. For the remaining datasets, we reserve a 10\% stratified sample for testing. In the following paragraphs we will be discussing some interesting datasets.

The only corpus in our collection that is truly multi-labeled is the iSarcasm dataset. The curators of iSarcasm created the collection by recruiting Twitter users and asking them to specify one sarcastic and three non-sarcastic tweets from their posted messages. Then, they asked the participants to provide a literal rephrase for every sarcastic message that conveys the same meaning. Furthermore, for every sarcastic message, the authors perform a second annotation stage where they further label these messages with irony, overstatement (hyperbole), understatement, satire, and rhetorical questions. In our work, we assume that the rephrases provided by the original participants are indeed literal sentences, however, we do not make the same assumption for the non-sarcastic messages that were also provided. In addition, since in our research we focus on six figurative language types, we ignore labels that are outside of this set. In such cases, we retain the sentence with only the sarcasm / not\_sarcasm, irony / not\_irony or hyperbole / not\_hyperbole labels.

The Sarcasm Corpus is a multi-class dataset centered around the binary classification task of sarcastic sentences. However, an extension of the dataset (separate file) contains sarcastic and non-sarcastic sentences that all have hyperbole. Since this was an addition to the main corpus, we cannot assume that the remaining files are completely devoid of hyperbole. Therefore, the hyperbolic sentences also have sarcasm / not\_sarcasm labels, but not the other way around. 

\begin{table*}[htb]
  \centering
  \small
  \begin{tabular}{>{\raggedright}p{0.1\linewidth}|c|c|c|c|c|c|c|c}
   \textbf{Datasets} & \textbf{Meta.}  & \textbf{Sarc.} & \textbf{Hyp.} & \textbf{Irony} & \textbf{Not Meta.} & \textbf{Not Sarc.} & \textbf{Not Hyp.} & \textbf{Not Irony} \\
    \hline
    Reddit Irony Corpus  & - 		 & - 		& - 	   & 483/54   & -		 & -         & -		 & 1271/141   \\ \hline
    Irony SemEval18 	 & - 		 & - 		& -	       & 1901/311 & -		 & -         & - 		 & 1916/473   \\ \hline
    iSarcasm 			 & - 		 & 713/133  & 40/6     & 155/19   & -		 & 3622/1059 & 4295/1186 & 4180/1173  \\ \hline
    Sarcasm Corpus 	     & - 		 & 4223/470 & 1047/117 & - 		  & -		 & 4224/469  & -		 & - 		  \\ \hline
    LCC 			  	 & 2732/304  & - 		& - 	   & - 		  & 3972/441 & -		 & -		 & -		  \\ \hline
  \end{tabular}
  \caption{(Appendix) Class distribution between train/test sets for each dataset. These datasets do not have `literal' annotations.}
  \label{tab:multi_dataset_distribution_1}
\end{table*}

The FLUTE dataset is also multi-class dataset, which means that each example is either metaphor, simile, sarcasm, or idiom. Each figurative sentence is paired with the two literal paraphrases, one aligning with the actual meaning of the figurative sentence, and the other communicating the opposite meaning. For instance, the figurative sentence \textit{"After a glass of wine, he loosened up a bit"} will have a literal counterpart \textit{"After a glass of wine, he relaxed up a bit"} and the opposite paraphrase would be \textit{"After a glass of wine, he stressed up a bit"}.

PIE-English is another interesting dataset. Here, the authors have automatically created a collection of sentences that contain possible idiomatic expressions. With further manual annotation efforts, they annotated each sentence whether its literal, therefore not idiomatic, or the idiom is constructed using euphemism, metaphor, personification, simile, parallelism, paradox, hyperbole oxymoron, or irony. Thus, every figurative sentence is an idiom plus an other figurative language class. In this work we focus on six figurative language types, so we ignore labels that are outside of this set. In such cases, we retain the sentence with only the idiom label.

\begin{table*}[htb]
  \centering
  \small
  \begin{tabular}{>{\raggedright}p{0.1\linewidth}|c|c|c|c|c|c|c}
   \textbf{Datasets}       & \textbf{Metaphor} & \textbf{Simile} & \textbf{Sarcasm} & \textbf{Hyperbole} & \textbf{Idiom} & \textbf{Irony}  & \textbf{Literal} \\
    \hline  
    MOVER 				   & -            & - 		   & - 	        & 906 / 101  & - 		    & - 	  & 1997 / 222   \\ \hline
    HypoGen 			   & -            & - 		   & - 	        & 1688 / 188 & - 		    & - 	  & 2585 / 287   \\ \hline
    EPIE 				   & -            & - 		   & - 	        & - 	     & 2485 / 276   & - 	  & 337 / 38     \\ \hline
    PIE-English 		   & 11330 / 1260 & 965 / 107  & - 	        & 41 / 5 	 & 12363 / 1375 & 27 / 3  & 966 / 107    \\ \hline
    FLUTE 				   & 625 / 124    & 625 / 125  & 2216 / 461 & - 	     & 884 / 125    & - 	  & 6368 / 1326  \\ \hline
    MSD23 				   & -            & 3218 / 358 & - 	        & - 	     & - 		    & - 	  & 4113 / 457   \\ \hline
    Figurative Comparisons & -            & 404 / 45   & - 	        & - 	     & - 		    & -       & 856 / 95     \\ \hline
    MOH 				   & 369 / 41     & - 		   & - 	        & - 	     & - 		    & - 	  & 1106 / 123   \\ \hline
  \end{tabular}
  \caption{(Appendix) Class distribution between train/test sets for each dataset. These datasets have `literal' annotations.}
  \label{tab:multi_dataset_distribution_2}
\end{table*}

\subsection{Appendix: Figurative Language Classification Binary Training Sets}
To train specilized binary models to detect FL features, we merge datasets annotated with examples relevant to each specific feature. For instance, to train a classifier for metaphors, we aggregate data from PIE-English, FLUTE, LCC, and MOH datasets. Similarly, for simile classification, we gather data from PIE-English, FLUTE, MSD23, and Figurative Comparisons datasets. Table \ref{tab:bin_training_distribution_appendix} shows the number of positive, negative and literal examples used to train each binary classifier.

\begin{table*}[htb]
  \centering
  \small
  \begin{tabular}{c|c|c|c}
    \textbf{Classifier} & \textbf{Positive} & \textbf{Negative} & \textbf{Literal} \\ \hline  
    Metaphor & 15056 & 3972 & 11084 \\ \hline
    Simile   & 5212  & 0    & 5212  \\ \hline
    Sarcasm  & 7152  & 3576 & 3576  \\ \hline
    Hyperbole & 3576 & 1861 & 1861  \\ \hline
    Idiom    & 15732 & 0    & 15732 \\ \hline
    Irony    & 2566  & 1283 & 1283  \\ \hline
  \end{tabular}
  \caption{(Appendix) The number of positive, negative and literal examples used to train each binary classifier.}
  \label{tab:bin_training_distribution_appendix}
\end{table*}

\subsection{Appendix: Figurative Language Classification Multi-label Training Set}
We use the fine-tuned specialized binary classification models to automatically tag our training corpora in a multi-label format. Table \ref{tab:multi_labe_dataset_appendix} shows the number of examples per figurative language class, as predicted by the binary classifiers. This dataset forms the basis of training our multi-task model. At a later step, this dataset gets split in train/dev set, where a 10\% stratified sample is reserved for development.

\begin{table*}[htb]
  \centering
  \small
  \begin{tabular}{c|c|c|c|c|c}
    \textbf{Metaphor} & \textbf{Simile} & \textbf{Sarcasm} & \textbf{Hyperbole} & \textbf{Idiom} & \textbf{Irony} \\ \hline  
    18981 & 6618 & 9906 & 13699 & 18604 & 10176 \\ \hline
  \end{tabular}
  \caption{(Appendix) Class distribution for the combined multi-labeled training dataset.}
  \label{tab:multi_labe_dataset_appendix}
\end{table*}

\subsection{Appendix: Figurative Language Classification Error Analysis}
In this subsection of the appendix, we present additional randomly selected examples where the model MFLM and binary model predictions do not align with human annotations. These additional examples are presented in Table \ref{tab:multi_analysis_appendix}. 

\begin{table*}[htb]
  \centering
  \small
  \begin{tabular}{c|c|c}
    \multicolumn{3}{c}{MFLM \& Binary models disagree with GT} \\
    \hline
    \textbf{GT}   & Literal &  \multirow{3}{*}{\begin{tabular}{p{0.6\linewidth}} \textit{Stupidity was \textbf{as important as} intelligence, and as difficult to attain.} \end{tabular}}\\
    \cline{1-2}
    \textbf{MFLM} & Simile & \\
    \cline{1-2}
    \textbf{Bin}  & Simile & \\
    \hline
    \hline
    \textbf{GT}   & Literal &  \multirow{3}{*}{\begin{tabular}{p{0.6\linewidth}} \textit{This office is \textbf{as lively as} a bustling beehive.} \end{tabular}}\\
    \cline{1-2}
    \textbf{MFLM} & Simile, Hyperbole & \\
    \cline{1-2}
    \textbf{Bin}  & Simile, Hyperbole & \\
    \hline
    \hline
    \textbf{GT}   & Literal &  \multirow{3}{*}{\begin{tabular}{p{0.6\linewidth}} \textit{They decided to continue, but within five minutes Sustad broke an ice hammer, \textbf{forcing them to retreat in mockingly perfect weather}.} \end{tabular}}\\
    \cline{1-2}
    \textbf{MFLM} & Sarcasm, Hyperbole & \\
    \cline{1-2}
    \textbf{Bin}  & Sarcasm, Hyperbole & \\
    \hline
    \hline
    \textbf{GT}   & Not Irony &  \multirow{3}{*}{\begin{tabular}{p{0.6\linewidth}} \textit{The letter and article seem to speak more of John Boehner wanting to \textbf{fire a gut} for criticizing the Pope.  Misleading title.} \end{tabular}}\\
    \cline{1-2}
    \textbf{MFLM} & Idiom, Sarcasm, Irony & \\
    \cline{1-2}
    \textbf{Bin}  & Idiom, Irony & \\
    
    \hline
    \multicolumn{3}{c}{ } \\
    \multicolumn{3}{c}{MFLM disagrees with GT, Binary model agrees with GT} \\
    \hline
    \textbf{GT}   & Literal &  \multirow{2}{*}{\begin{tabular}{p{0.6\linewidth}} \textit{\textbf{I ace through} the work.} \end{tabular}}\\
    \cline{1-2}
    \textbf{MFLM} & Metaphor & \\
    \hline
    \hline
    \textbf{GT}   & Not Sarcasm, Not Irony &  \multirow{2}{*}{\begin{tabular}{p{0.6\linewidth}} \tiny{\textit{\textbf{Full throttle}? 11 players changed and playing the philosophy that the manager wants isn't grounds for slagging! Especially when we win! \textbf{Clutching at straws} here!}} \end{tabular} } \\
    \cline{1-2}
    \textbf{MFLM} & Metaphor, Idiom, Irony & \\
    \hline
    \hline
    \textbf{GT}   & Literal &  \multirow{2}{*}{\begin{tabular}{p{0.6\linewidth}} \textit{This \textbf{dirty money} we're using to finance the campaign is a risk!} \end{tabular}} \\
    \cline{1-2}
    \textbf{MFLM} & Metaphor & \\
    \hline
    \hline
    \textbf{GT}   & Literal &  \multirow{2}{*}{\begin{tabular}{p{0.6\linewidth}} \textit{The leaves clog our drains in the Fall} \end{tabular}} \\
    \cline{1-2}
    \textbf{MFLM} & Metaphor & \\
    \hline
    \hline
    \textbf{GT}   & Not Metaphor &  \multirow{2}{*}{\begin{tabular}{p{0.6\linewidth}} \textit{If you are trying to claim gun control is not incremental I am first going to \textbf{laugh my head off} at such an obviously stupid statement.} \end{tabular}} \\
    \cline{1-2}
    \textbf{MFLM} & Metaphor, Irony, Hyperbole & \\
    \hline
    
  \end{tabular}
  \caption{(Appendix) Samples where model predictions do not align with human annotations.}
  \label{tab:multi_analysis_appendix}
\end{table*}

\subsection{Appendix: Authorship Attribution Baselines}
In this section of the appendix we provide further details regarding the Stylometric features of our Authorship Attribution (AA) baseline approach. We implement 52 text metrics using the cophi\footnote{cophi: \url{https://github.com/cophi-wue/cophi-toolbox}} and textstat\footnote{textstat: \url{https://github.com/textstat/textstat}} Python packages. These metrics are used to form a document vector with 52 stylometric features. In the Table \ref{tab:aa_stylo_features_appendix} we list the feature names along with implementation notes.

\begin{table*}[htb]
  \centering
  \small
  \begin{tabular}{l|p{0.65\linewidth}}
    \hline
    \textbf{Feature Name}               & \textbf{Notes} \\ \hline
    Average Word Length Chars			& Average number of characters per word. \\ \hline
    Average Syllables Per Word 			& Average number of syllables per word. \\ \hline
    Average Sentence Length 			& Average number of words per sentence. \\ \hline
    Average Sentence Length Chars 		& Average number of characters per sentence. \\ \hline
    Average Word Frequency Class 		& \citep{meyer2007plagiarism} \\ \hline
    Type Token Ratio 					& Number of unique words (types) over the total number of words (tokens). \\ \hline
    Digit Ratio 			            & Number of numerical characters over total number of characters. \\ \hline
    Puncuations Ratio 					& Number of punctuation characters over total number of characters. \\ \hline
    Uppercase Ratio 					& Number of uppercase letter characters over total number of characters. \\ \hline
    Special Characters Ratio 			& Number of special characters over total number of characters. \\ \hline
    Stopword Ratio 						& Number of stopwords over total number of words. \\ \hline
    Functional Words Ratio 				& Number of functional words over total number of words. \\ \hline
    Hapax Legomena Ratio 				& Number of words that appear once over total number of words. \\ \hline
    Hapax Dislegomena Ratio 			& Number of words that appear twice over total number of words. \\ \hline
    Automated Readability Metric 		& \citep{senter1967automated} \\ \hline
    Flesch Reading Ease Metric 			& \citep{kincaid1975derivation} \\ \hline
    Flesch Kincaid Grade Metric 		& \citep{kincaid1975derivation}  \\ \hline
    Dale Chall Readability Metric 		& \citep{dale1948formula}  \\ \hline
    New Dale Chall Readability Metric 	& \citep{chall1995readability}  \\ \hline
    Spache Readability Metric 			& \citep{spache1953new}  \\ \hline
    Gunning Fog Metric 					& \citep{gunning1952technique}  \\ \hline
    Lix Index 							& Average sentence length plus the percentage of words of more than six letters. \\ \hline
    Rix Index 							& \citep{anderson1981analysing}  \\ \hline
    Fernandez Huerta Index 			    & \citep{fernandez1959medidas}  \\ \hline
    Szigriszt Pazos Index 				& \citep{pazos1993sistemas}  \\ \hline
    Crawford Index 					    & \citep{crawford1985formula} \\ \hline
    Mcalpine Eflaw Metric 				& \citep{mcalpine2012plain} \\ \hline
    Guiraud R Metric 					& \citep{guiraud1954caractères} \\ \hline
    Herdan C Metric 					& \citep{herdan2006definite} \\ \hline
    Dugast K Metric 					& \citep{dugast1979vocabulaire} \\ \hline
    Maas A2 Metric 						& \citep{mass1972zusammenhang} \\ \hline
    Dugast U Metric 					& \citep{dugast1980statistique} \\ \hline
    Tuldava LN Metric 					& \citep{tuldava1977quantitative} \\ \hline
    Brunet W Metric 					& \cite{brunet1978vocabulaire} \\ \hline
    Corrected Token Type Ratio 			& \citep{carroll1964language} \\ \hline
    Summer S Index 					    & Similar to TTR, $S = log(log(\text{types}))/log(log(\text{tokens}))$. \\ \hline
    Sichel S Metric 					& \citep{sichel1975distribution} \\ \hline
    Michea M Metric 					& \citep{michea1969repetition, hichea1971relation} \\ \hline
    Honore H Metric 					& \citep{honore1979some} \\ \hline
    Shannon Entropy 		            & \citep{shannon1948mathematical} \\ \hline
    Yule K Metric 						& \citep{yule2014statistical} \\ \hline
    Simpson D Metric 					& \citep{simpson1949measurement} \\ \hline
    Herdan VM Metric 					& \citep{herdan1955new} \\ \hline
    Coleman Liau Metric 				& \citep{coleman1975computer} \\ \hline
    Linsear Write Metric 				& \citep{o1966gobbledygook} \\ \hline
    Smog Metric 						& \citep{mc1969smog} \\ \hline
    Threshold Word Length H Ratio 		& Number of words with more than 5 characters over total number of words. \\ \hline
    Threshold Word Length L Ratio 		& Number of words with less than 5 characters over total number of words. \\ \hline
    Threshold Syllables Per Word H Ratio & Number of words with more than 2 syllables over total number of words. \\ \hline
    Threshold Syllables Per Word L Ratio & Number of words with less than 2 syllables over total number of words. \\ \hline
    Threshold Sentence Length H Ratio 	 & Number of sentences with more than 17 words over the total number or sentences. \\ \hline
    Threshold Sentence Length L Ratio 	 & Number of sentences with less than 17 words over the total number or sentences. \\ \hline
  \end{tabular}
  \caption{(Appendix) List of Stylometric feature names and implementation notes.}
  \label{tab:aa_stylo_features_appendix}
\end{table*}

\end{document}